# A Supervised-Learning-Based Strategy for Optimal Demand Response of an HVAC System

Young-Jin Kim, *Member, IEEE*

*Abstract*—The large thermal capacity of buildings enables heating, ventilating, and air-conditioning (HVAC) systems to be exploited as demand response (DR) resources. Optimal DR of HVAC units is challenging, particularly for multi-zone buildings, because this requires detailed physics-based models of zonal temperature variations for HVAC system operation and building thermal conditions. This paper proposes a new strategy for optimal DR of an HVAC system in a multi-zone building, based on supervised learning (SL). Artificial neural networks (ANNs) are trained with data obtained under normal building operating conditions. The ANNs are replicated using piecewise linear equations, which are explicitly integrated into an optimal scheduling problem for price-based DR. The optimization problem is solved for various electricity prices and building thermal conditions. The solutions are further used to train a deep neural network (DNN) to directly determine the optimal DR schedule, referred to here as supervised-learning-aided meta-prediction (SLAMP). Case studies are performed using three different methods: explicit ANN replication (EAR), SLAMP, and physics-based modeling. The case study results verify the effectiveness of the proposed SL-based strategy, in terms of both practical applicability and computational time, while also ensuring the thermal comfort of occupants and cost-effective operation of the HVAC system.

*Index Terms*—artificial neural networks (ANNs), demand response (DR), heating, ventilating, and air-conditioning (HVAC), multi-zone building, supervised-learning-aided meta-prediction

## I. INTRODUCTION

The commercial building sector accounted for 35% of the electricity consumption in the United States in 2010 [1]. Heating, ventilating, and air-conditioning (HVAC) loads represent about 30% of the electricity usage in a commercial building [2]. Therefore, distribution system operators (DSOs) have implemented various demand response (DR) programs to adjust the power inputs of HVAC units and thereby improve the operational efficiency and stability of power grids [3], [4]. Meanwhile, compressors and fans have evolved from single- to variable-speed units, increasing the operating flexibility and energy efficiency of HVAC loads. Building energy management systems (BEMSs) are also increasingly updated via internet-of-things (IoT) sensing, enhanced communication, and big data analysis. The development of such technologies enables DSOs and DR aggregators to further improve DR programs.

For optimal DR of HVAC systems, model predictive control (MPC) methods were widely adopted in previous studies (e.g., [5] and [6]), using physics-based analyses of HVAC system operations and building thermal dynamics. Such analyses require nonlinear models of sub-devices in HVAC systems. These models include numerous parameters, most of which are extracted using parameter estimation techniques. Moreover, the indoor temperatures in a multi-zone building are affected by heat transfer across thermal zones and heat gains from the HVAC equipment and building environment, which vary widely by building type, size, and structure. The time-consuming process of modeling and parameter tuning makes it difficult to apply physics-model-based DR strategies to various buildings with different types of HVAC systems.

To overcome this challenge, machine learning (ML) has been considered in recent papers, as listed in Table I. Specifically, 10 criteria [i.e., (a)–(j)] have been chosen to compare similarities and differences among the papers. For example, reinforcement learning (RL) was adopted in [7]–[9]; however, only on-off operations or discrete power inputs of HVAC systems were considered, as significant computational time is required to search for an optimal DR policy over large state and action spaces. Moreover, during the training process, HVAC systems should operate randomly to expand state and action spaces, and consequently to better define the reward functions. Nonlinear models of building thermal dynamics also need to be developed and verified using actual data collected under various conditions of HVAC system operation and building thermal environment. This makes it difficult to apply RL-based DR strategies in practice, as in the case involving physics-model-based strategies. Therefore, supervised learning (SL) was used in [10]–[17], in which various types of artificial neural networks (ANNs) were implemented and trained to model HVAC system operations, considering the trade-off between modeling accuracy and computational time. Using the trained ANNs, scheduling problems were subsequently formulated to ensure the optimal operation of HVAC systems for the next 24-h time period.

In most papers, the trained ANNs were treated as black-box models or represented as functional equations, so they required iterative, heuristic algorithms (e.g., firefly, GA, and PSO) to find solutions. For the algorithms, solutions often fall into numerous local optima. Moreover, the scheduling problems need to be solved iteratively to select the best local optimal solution among those obtained so far, which ultimately increases the computational time [7]. On the other hand, in [11] and [15], trained ANNs were replicated using a set of nonlinear equations, which enabled the application of non-iterative, deterministic nonlinear solvers; however, the nonlinear solvers in [11] and [15] still could not guarantee the global optimality of the solutions.

With regard to objective functions, the optimal DR of HVAC systems for time-varying electricity prices and corresponding cost savings were analyzed in [10], [11], and [16], whereas energy savings were the principal focus of the other papers. Even for price-based DR, time-of-use (TOU) rates were commonly applied as the electricity price profiles. However, increasing attention has been devoted recently to real-time pricing (RTP) schemes [18], [19], wherein prices change on an hourly basis, depending on the load demand variation in the distribution grid. Therefore, the HVAC load can be more flexibly shifted or curtailed during on-peak hours, improving the effectiveness of DR programs; the RTP schemes were not reflected in [10]–[17]. In [7]–[17], the optimal HVAC load schedules were compared with current non-DR profiles or pre-determined with fixed temperature set-points. Therefore, for conservative evaluation of the SL-based methods, the optimal schedules need to be verified through a comparison with those for an ideal MPC method.

Based on these observations, this paper proposes a new SL-based strategy for optimal DR of an HVAC system in a multi-zone commercial building. Specifically, continuous variations in the power inputs of the HVAC system are scheduled by reference to hourly varying electricity prices, while maintaining indoor temperatures within acceptable ranges. An ANN with a feedback loop, time-delayed inputs, and multiple hidden layers is implemented to estimate the temperature variation of each zone for change in the power input of the HVAC system. The ANN is then trained with data obtained under normal operating conditions of the building and explicitly replicated using a set of piecewise linear equations; this proposed approach is termed an explicit ANN replication (EAR). For optimal DR scheduling, it allows formulation of an optimization problem using only mixed-integer linear equations, unlike [10]–[17]. The boundaries of the feasible solution area are extended and set by the mixed-integer linear constraints; therefore, a linear, deterministic solver (e.g., MILP) can readily be applied, ensuring global optimality of the solution. The EAR method is further exploited to mitigate the requirement of DR aggregators or end-users to solve the optimization problem on a daily basis. Using the EAR method, the optimal solutions are obtained offline for historical data on electricity prices and building thermal conditions and for training a deep neural network (DNN). The optimal DR schedule can then be obtained directly as the outputs of the trained DNN, rather than by solving the optimization problem, for the forecasted prices and thermal conditions during the next 24 h. This proposed technique is termed *supervised-learning-aided meta-prediction* (SLAMP), which enables a significant reduction in the computational time for optimal DR scheduling. Simulation case studies are performed using three methods: the proposed EAR and SLAMP methods and the conventional fully informed physics-model-based method. The case study results demonstrate that the proposed SL-based methods are effective in terms of both practical applicability and computational time, while also ensuring the thermal comfort of occupants and cost-effective operation of the HVAC system.

Compared to [7]–[17], the original contributions of this paper are summarized as follows:

• For the EAR method, a comprehensive set of piecewise linear equations is presented to explicitly replicate a general type of ANN with feedback loops, time-delayed inputs, and multiple hidden layers, which can be directly integrated as constraints into an optimal DR scheduling problem. This enables the optimization problem to be solved using an off-the-shelf MILP solver, ensuring global optimality of the solution within a reasonable time and easy incorporation into practical DR programs [20].

• The SLAMP method, developed using the EAR method, can directly predict the optimal schedule of HVAC system operation for the next 24 h using only the forecasted electricity prices and building thermal conditions. This significantly reduces the computational time for optimal DR scheduling, as the optimization problem solving is replaced with DNN-based prediction. The SLAMP method can assist the DSO or DR aggregators in identifying the optimal price sensitivities of actual, individual HVAC systems better than simplified

TABLE I. PREVIOUS STUDIES ON THE APPLICATION OF MACHINE LEARNING ALGORITHMS TO OPTIMAL SCHEDULING OF HVAC SYSTEM OPERATION

| Ref. | (a) ML type | (b) HVAC operation | (c) ANN type | (d) Building type | (e) ANN integration | (f) Objective function ||||| (g) Electricity pricing | (h) Optimization algorithm | (i) Scheduling comparison | (j) SLAMP for optimal DR |
|------|------|------|--------|------|------|------|------|------|------|------|-------|------|------|
| | | | | | | CS | ES | TM | Others | | | | |
| Proposed | SL | C | D-NARX | MZ | ER | ○ | | ○ | | RTP | MILP | MPC | ○ |
| [7] | RL | O | DNN | SZ | - | ○ | | | PL | TOU | DQL, DPG | CO | |
| [8] | RL | D | MDP | SZ | - | ○ | | ○ | | TOU | DQL | CO | |
| [9] | RL | D | DQN | MZ | - | ○ | | ○ | | TOU | DQL | PSS | |
| [10] | SL | O | MLP | SZ | BB | ○ | | | | TOU | nonlinear | PSS | |
| [11] | SL | C | NARX | SZ | ER | ○ | | ○ | | TOU | MINLP | PSS | |
| [12] | SL | C | NARX | SZ | BB | | ○ | ○ | | - | PSO | CO | |
| [13] | SL | O | MLP | MZ | BB | | ○ | | | - | GA | PSS | |
| [14] | SL | C | MLP | MZ | BB | | | ○ | TR | - | EA, PSO, HS | PSS | |
| [15] | SL | O | D-NARX | MZ | ER | | | | TD | TOU | BB | PSS | |
| [16] | SL | C | D-NARX | MZ | BB | ○ | ○ | ○ | | TOU | GA | PSS | |
| [17] | SL | C | RNN | MZ | BB | | ○ | ○ | | - | MGD | CO | |

(b) **C**: continuous input power variation, **D**: discrete input power variation, **O**: on-off operation.
(c) **D-NARX**: deep NARX, **DNN**: deep neural network, **DQN**: deep Q-network, **MDP**: Markov decision process, **MLP**: multi-layer perceptron, **NARX**: nonlinear auto-regressive network with exogenous inputs, **RNN**: recurrent neural network.
(d) **MZ**: multi-zone building, **SZ**: single-zone (or identical-zone) building.
(e) **ER**: explicit replication, **BB**: black-box modeling (or functional expression).
(f) **CS**: operating cost saving, **ES**: energy saving, **PL**: peak-load reduction, **TD**: difference between the temperature set-point and the actual (or ambient) temperature, **TR**: ramp rate of indoor temperature, **TM**: indoor temperature maintenance.
(g) **RTP**: real-time (or hourly varying) pricing, **TOU**: time-of-use pricing.
(h) **BB**: branch and bound, **DPG**: deep policy gradient, **DQL**: deep Q-learning, **EA**: evolutionary algorithm, **GA**: genetic algorithm, **HS**: harmony search, **MGD**: momentum gradient descent, **MILP**: mixed integer linear programming, **MINLP**: mixed integer nonlinear programming, **PSO**: particle swarm optimization.
(i) **CO**: currently observed profiles, **PSS**: pre-determined (or fixed) set-points, **MPC**: physics-based model predictive control.

price-and-quantity curves [21].
- We present and incorporate a procedure to select the ANN and DNN architectures for the least over-fitting into the EAR and SLAMP methods, respectively, to enhance the generalization capability of the networks in reflecting the building thermal responses to HVAC system operations, and hence to improve the performances of the proposed SL-based methods.
- Case studies are performed considering the RTP scheme and the fully informed physics-model-based method. The case study results demonstrate the effectiveness of the proposed SL-based methods in shifting the HVAC load with RTP rate variation and in reducing the operating costs of the HVAC system. In particular, a comparison with the physics-based method enables conservative evaluation of the performances of the proposed methods.

## II. SUPERVISED LEARNING OF THE THERMAL RESPONSE OF A MULTI-ZONE BUILDING TO HVAC SYSTEM OPERATION

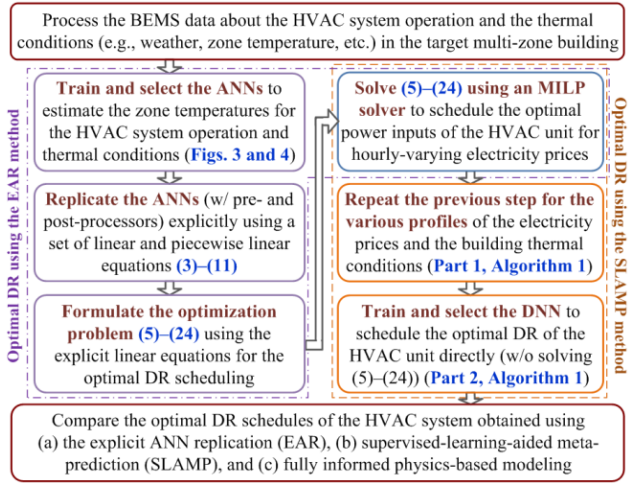

Fig. 1. Overall flowchart of the optimal DR scheduling strategy using the proposed EAR and SLAMP methods.

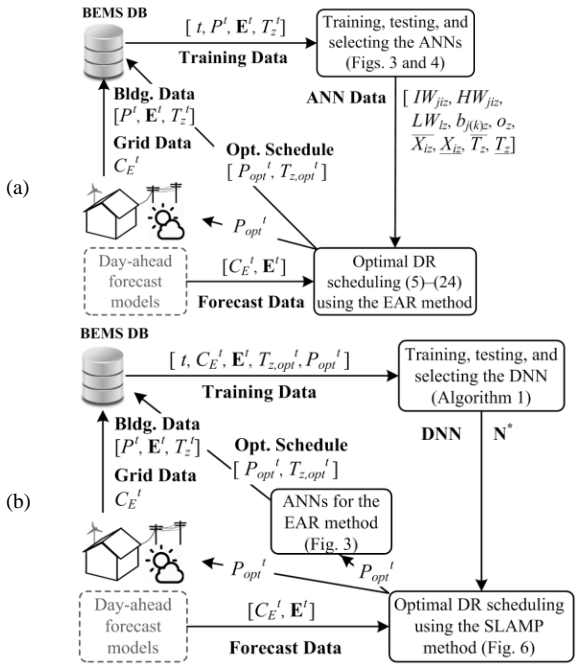

Fig. 2. Schematic diagrams representing the input and output data flows for the proposed (a) EAR and (b) SLAMP methods.

Fig. 1 presents the overall framework for the proposed SL-based DR strategy, featuring three main parts that (a) develop the ANNs to model the HVAC system operation and building thermal environment, (b) formulate and solve the optimization problem using the EAR method, and (c) develop the SLAMP method using the optimal solutions for direct determination of the optimal DR schedule. Fig. 2(a) and (b) represent input and output data flows for the proposed EAR and SLAMP methods, respectively, in which the neural networks are trained with building and power grid data. In the EAR method, the weighting coefficients and bias values in the trained ANNs are used to formulate the optimal DR scheduling problem. In the SLAMP method, the trained DNN is used to determine the optimal power inputs of the HVAC system and the corresponding indoor temperatures in conjunction with the ANNs. As this paper focuses on DR service provision via the HVAC load scheduling for the following 24-h time period [22], the proposed DR strategy needs to be integrated with the forecast models of the electricity prices and building thermal conditions. Various forecast models were studied extensively, as described in [23] and [24]; thus, the historical data records were used in this study for simplicity. The optimal scheduling data are stored with the building and power grid data in BEMS databases. Each block in Figs. 1 and 2 will be further explained in Sections II and III.

### A. ANN architecture

The indoor temperature $T_z^t$ of each zone $z$ at time $t$ is affected by the power inputs $P^t$ of the HVAC system and the thermal conditions $\mathbf{E}^t$ of the multi-zone building during the time period from $t-\tau$ to $t$. Moreover, for the same values of $P^t$ and $\mathbf{E}^t$, $T_z^t$ differs by $T_z^{(t-1)}$. In other words, $T_z^t$ serves as both the state and output variables in a state-space model of building thermal dynamics. This implies that an ANN with time-delayed inputs of $P^t$ and $\mathbf{E}^t$ and feedback loops of $T_z^t$ can appropriately model the building thermal dynamics. In this paper, a direct method to control the HVAC system is considered [25], [26] while taking advantage of a variable speed drive (VSD). For the VSD-interfaced HVAC system, the reference power inputs $P_{set}^t$ can be determined to maintain $T_z^t$ within an acceptable range for time-varying $\mathbf{E}^t$ and sent to the VSD via the communication system of the BEMS, so that $P^t$ can follow $P_{set}^t$. Due to the fast time response of the VSD, $P^t$ is similar to $P_{set}^t$. In contrast, for an indirect control method, the temperature set points $T_{z,set}^t$ are adjusted first to change $P_{set}^t$ and then $T_z^t$. Therefore, the number of ANN input variables for the direct method is smaller than for the indirect method; i.e., [$t$, $P^t$, $\cdots$, $P^{t-\tau1}$, $\mathbf{E}^t$, $\cdots$, $\mathbf{E}^{t-\tau2}$, $T_z^{t-1}$, $\cdots$, $T_z^{t-\tau3}$] and [$t$, $T_{z,set}^t$, $\cdots$, $T_{z,set}^{t-\tau1}$, $P^{t-1}$, $\cdots$, $P^{t-\tau2}$, $\mathbf{E}^t$, $\cdots$, $\mathbf{E}^{t-\tau3}$, $T_z^{t-1}$, $\cdots$, $T_z^{t-\tau4}$], respectively. Consequently, the optimal scheduling problem, discussed in Section III-A, can be formulated more simply in the direct method than in the indirect method. Moreover, the direct method can readily be integrated into practical DR programs in which the DSO makes contracts on DR capacity in kW [27]. Note that the proposed DR strategy also can be implemented using the indirect method with slight modifications to the ANN input variables.

To further improve modeling accuracy, multiple hidden

-3-

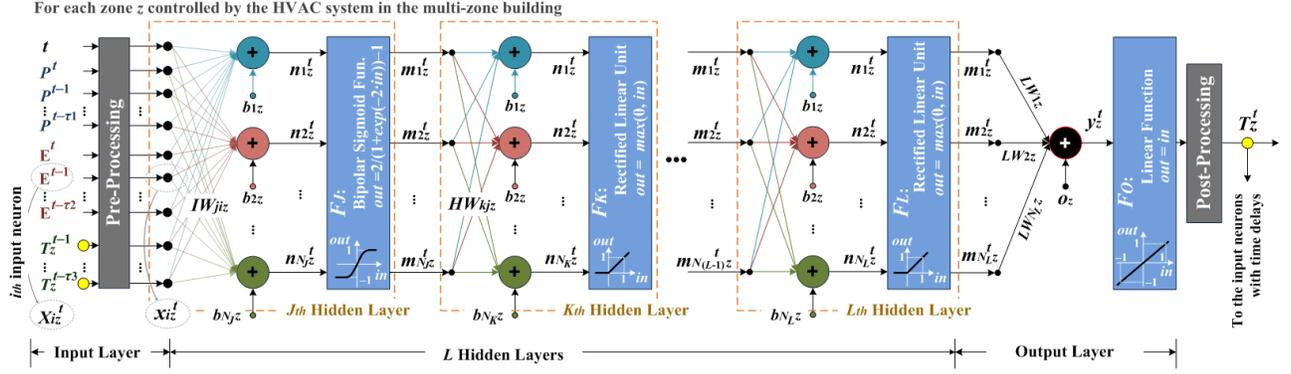

Fig. 3. ANN model estimating the indoor temperature of zone $z$ for the HVAC system operation $P^t$ and the building thermal conditions $\mathbf{E}^t$.

layers are included, wherein the sigmoid functions or rectified linear units (ReLUs) are commonly used as activation functions [28], [29]. Therefore, as shown in Fig. 3, the ANN has been implemented in the form of a deep nonlinear auto-regressive network with exogenous inputs (D-NARX). Note that different multi-layer networks can also be applied to the proposed DR strategy, with minor modifications.

The ANN also includes a pre-processor to normalize the input dataset $\mathbf{X}_z^t$; otherwise, the gradient of a training algorithm can be very small, slowing the ANN training process. Based on the availability of building operating data in the BEMS, $\mathbf{E}^t$ can include ambient temperature, solar insolation, wind speed, humidity, building thermal load, or occupancy rate [13]–[17], [30], [31], the values of which vary over different ranges. A post-processor is then required to reverse-transform the normalized indoor temperature $y_z^t$ into the same unit as the original temperature $T_z^t$.

### B. ANN training and architecture parameter selection

The ANN, shown in Fig. 3, is trained with historical building data using the *trainscg* function provided in MATLAB for the application of scaled conjugate gradient backpropagation [32]. During the training, the weighting coefficients (i.e., $IW_{jiz}$, $HW_{kjz}$, and $LW_{lz}$) and bias values (i.e., $b_{j(k)z}$ and $o_z$) are determined for all input, hidden, and output neurons. The ANN features different weighting and bias values for each thermal zone $z$. Considering the ANN complexity, the values are set to be constant during the scheduling period $1\ h \leq t \leq N_H$. Moreover, the ANN architecture, such as the activation functions, the input variables (apart from $T_z^{(t-\tau)}$), and the number of hidden layers and neurons, is the same for all zones. The training and test performances of the ANN for each zone can be estimated using the normalized mean squared errors (NMSEs) [10] as

$$e(\mathbf{T}_z, \mathbf{T}_z') = 1 - \frac{\sqrt{\sum_{t=1}^{N_T}\left(T_z^t - T_z^{t\prime}\right)^2}}{\sqrt{\sum_{t=1}^{N_T}\left(T_z^t - T_{z,avg}\right)^2}} \quad \text{where } T_{z,avg} = \frac{1}{N_T}\sum_{t=1}^{N_T} T_z^t, \quad (1)$$

where $T_z^{t\prime}$ is the predicted value of $T_z^t$ and $N_T$ is the number of training (or test) data points.

In various applications of ANNs, over-fitting is a common issue, particularly when the size and variability of training data are limited. Over-fitting can be lessened or even avoided using different network architectures [33]–[35], for example, with regard to connections between hidden neurons, number of hidden neurons, and types of activation functions. Therefore, a rather simple procedure is established in this study to select the architecture with which the ANN is the least over-fitted, as shown in Fig. 4. Specifically, the parameter set is defined as $\mathbf{Z} = [\boldsymbol{\tau}, G, \mathbf{U}, \mathbf{F}]$, where $\boldsymbol{\tau}$ is the set of $\tau_1$, $\tau_2$, and $\tau_3$, which are the largest time delays for $P^t$, $\mathbf{E}^t$, and $T_z^t$, respectively, as shown in Fig. 3. $G$ is the number of hidden layers, and $\mathbf{U}$ is the set of $U_g$, which is the number of neurons for the $g$th hidden layer. Similarly, $\mathbf{F}$ is the set of $F_g$, which is the type of activation function for the $g$th hidden layer. The search range for the ANN architecture selection is then defined by the minimum and maximum values of $\boldsymbol{\tau}$, $G$, and $U$, as well as the number of activation function types, as shown in Table II.

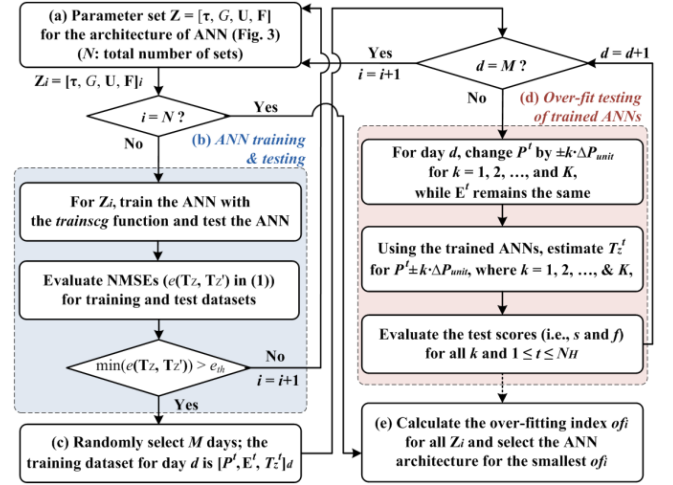

Fig. 4. Schematic diagram of the procedure to select the parameter set for the architecture with which the ANN is the least over-fitted.

TABLE II. SEARCH RANGE FOR THE PARAMETER SET $\mathbf{Z}$ OF THE ANNS

| Parameter | $\tau_{1,max}$ | $\tau_{2,max}$ | $\tau_{3,max}$ | $\Delta\tau_1$ | $\Delta\tau_2$ | $\Delta\tau_3$ | $G_{max}$ | $\Delta G$ | $U_{g,max}$ | $U_{g,min}$ | $\Delta U_g$ | $F_g$ |
|---|---|---|---|---|---|---|---|---|---|---|---|---|
| Value | 4 | 4 | 4 | 1 | 1 | 1 | 4 | 1 | 25 | 5 | 5 | [†]sig. or ReLU |

([†]: sigmoid function)

As shown in Fig. 4(a), the ANN is trained and tested using $\mathbf{Z}_i$. The corresponding NMSEs are evaluated using (1). If the minimum value of the NMSEs for all zones is higher than $e_{th}$ = 0.9, $M$ = 10 days are randomly selected, as shown in Fig. 4(c). The day $d$ is characterized by $[P^t, \mathbf{E}^t, T_z^t]_d$ for $1\ h \leq t \leq N_H$ different from those for the other $M - 1$ days. Fig. 4(d) shows that for each $d$, $P^t$ gradually increases by $k\cdot\Delta P_{unit}$ from $k = 1$ to $K$; $P^t$ should be maintained between 0 and $P_{max}$. Using the trained ANN, $T_z^t$ is estimated for $P^t + k\cdot\Delta P_{unit}$, while $\mathbf{E}^t$ remains the same. At each time $t$, the value of $s = 0$ is assigned as the test score $sc_{t,k,d,i}$ for the case where the output $T_z^t$ of the trained ANN

-4-

for $P^t+k\cdot\Delta P_{unit}$ is smaller than $T_z^t$ for $P^t+(k-1)\cdot\Delta P_{unit}$. This implies that the ANN has succeeded in reflecting the building thermal response to the HVAC system operation: i.e., $T_z^t$ decreases when $P^t$ increases for the same $\mathbf{E}^t$. In contrast, $f = 5$ is assigned as $sc_{t,k,d,i}$ for the case where $T_z^t$ for $P^t+k\cdot\Delta P_{unit}$ is higher than $T_z^t$ for $P^t+(k-1)\cdot\Delta P_{unit}$. This indicates that the ANN has failed to reflect the thermal response. Analogously, $s$ is assigned as $sc_{t,k,d,i}$ when $T_z^t$ for $P^t-k\cdot\Delta P_{unit}$ is higher than $T_z^t$ for $P^t-(k-1)\cdot\Delta P_{unit}$; otherwise, $f$ is assigned. Note that because the data on $T_z^t$ for the condition on $P^t\pm k\cdot\Delta P_{unit}$ and $\mathbf{E}^t$ do not exist in the training datasets, it can be only known whether $T_z^t$ should increase or decrease for $P^t\pm k\cdot\Delta P_{unit}$. Therefore, the two discrete values $s$ and $f$ are used in the over-fit test.

Fig. 4(e) shows that for each $k$ on day $d$ for $\mathbf{Z}_i$, the average $avg_{k,d,i}$ and standard deviation $std_{k,d,i}$ of $sc_{t,k,d,i}$ for $1\text{ h}\leq t\leq N_H$ are estimated as

$$avg_{k,d,i} = \frac{1}{N_H}\sum_{t=1}^{N_H} sc_{t,k,d,i}, \qquad \forall k, \forall d, \forall i, \quad (2\text{-a})$$

$$std_{k,d,i} = \left(\frac{1}{N_H}\sum_{t=1}^{N_H} sc_{t,k,d,i}^2 - avg_{k,d,i}^2\right)^{1/2}, \quad \forall k, \forall d, \forall i. \quad (2\text{-b})$$

For all $k$ on day $d$, $avg_{d,i}$ and $std_{d,i}$ are then calculated as the average and standard deviation of $avg_{k,d,i}$ and $std_{k,d,i}$, respectively. Similarly, for all $d$, $avg_i$ and $std_i$ can be further estimated as the average and standard deviation of $avg_{d,i}$ and $std_{d,i}$, respectively. In this paper, the over-fitting index of the ANN is defined as $of_i = c_1\cdot avg_i + c_2\cdot std_i$. The ANN implemented with $\mathbf{Z}_i$, leading to the smallest value of $of_i$, is then selected for the optimization problem formulation, discussed in Section III-A. Both $c_1$ and $c_2$ are set to 0.5 to calculate $of_i$; different values can be used without loss of generality. Note that the more sophisticated methods discussed, for example, in [33]–[35] can also be used to better combat ANN over-fitting.

*C. Explicit replication of trained ANNs*

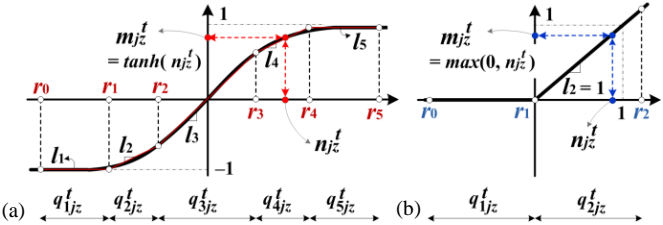

Fig. 5. Piecewise linear approximations of the activation functions: (a) the sigmoid function and (b) rectified linear units (ReLU).

The trained ANN, shown in Fig. 3, is explicitly replicated using a set of mixed-integer linear equations. Specifically, the output $n_{jz}^t$ of the hidden neuron $j \in \mathbf{H}_{1z}$ in the first hidden layer can be estimated as:

$$n_{jz}^t = \sum_{i=1}^{N_I} IW_{jiz}x_{iz}^t + b_{jz}, \qquad \forall j\in\mathbf{H}_{1z}, \forall t, \forall z, \quad (3)$$

where $x_{iz}^t$ are the normalized data of the $i_{\text{th}}$ input neuron at time $t$ for zone $z$. Similarly, the output $n_{kz}^t$ of the hidden neuron $k \in \mathbf{H}_{Kz}$ is calculated using the output $m_{jz}^t$ of the activation function for the hidden neuron $j \in \mathbf{H}_{Jz}$ for $J = K-1$, as:

$$n_{kz}^t = \sum_{j=1}^{N_J} HW_{kjz}m_{jz}^t + b_{kz}, \qquad \forall k\in\mathbf{H}_{Kz}, \forall t, \forall z. \quad (4)$$

In (4), $m_{jz}^t$ can be further expressed using $n_{jz}^t$ as

$$m_{jz}^t - \sum_{s=1}^{N_S} l_s q_{sjz}^t = F_{J,\min}, \qquad \forall j, \forall t, \forall z, \quad (5\text{-a})$$

$$n_{jz}^t = r_0 + \sum_{s=1}^{N_S} q_{sjz}^t, \qquad \forall j, \forall t, \forall z, \quad (5\text{-b})$$

$$-q_{1jz}^t + (r_1 - r_0)w_{1jz}^t \leq 0, \qquad \forall j, \forall t, \forall z, \quad (6\text{-a})$$

$$q_{1jz}^t \leq (r_1 - r_0), \qquad \forall j, \forall t, \forall z, \quad (6\text{-b})$$

$$-q_{sjz}^t + (r_s - r_{(s-1)})w_{sjz}^t \leq 0, \quad \forall s\in[2,\cdots,(N_S-1)], \forall j, \forall t, \forall z, \quad (7\text{-a})$$

$$q_{sjz}^t - (r_s - r_{(s-1)})w_{(s-1)jz}^t \leq 0, \quad \forall s\in[2,\cdots,(N_S-1)], \forall j, \forall t, \forall z, \quad (7\text{-b})$$

$$-q_{N_S jz}^t \leq 0, \qquad \forall j, \forall t, \forall z, \quad (8\text{-a})$$

$$q_{N_S jz}^t - (r_{N_S} - r_{(N_S-1)})w_{(N_S-1)jz}^t \leq 0, \qquad \forall j, \forall t, \forall z. \quad (8\text{-b})$$

In (5), $l_s$ is the piecewise linear gradient of $m_{jz}^t$, attributable to the linear block $s$ of $n_{jz}^t$, as shown in Fig. 5. $F_{J,min}$ is the minimum value of the activation function $F_J$, and $N_S$ is the number of piecewise linear blocks. In (6)–(8), $q_{sjz}^t$ is the value assigned in block $s$ of $n_{jz}^t$ and $w_{sjz}^t$ is the binary variable completing the piecewise linearization. For each hidden neuron, $F_J$ can be approximated using (5)–(8) by dividing the entire range of $n_{jz}^t$ into $N_S$ linear blocks, as shown in Fig. 5, where $r_0$ is an arbitrarily large negative number. For the sigmoid function, $N_S$ was set to 5, taking into account the trade-off between modeling precision and computational burden. The higher $N_S$, the higher the precision of the piecewise linear approximation. For the ReLU function, $N_S = 2$ enables exact piecewise linearization, as shown in Fig. 5(b). The piecewise linearization aims to explicitly replicate the trained ANNs using mixed-integer linear equations that can be integrated directly into the optimal DR scheduling problem, as discussed in Section III-A. It consequently enables MILP application to solve the optimization problem non-iteratively within a reasonable computational time and ensures the global optimality of the solution, unlike heuristic or nonlinear algorithms in [10]–[17].

The output neuron produces:

$$y_z^t = \sum_{l=1}^{N_L} LW_{lz}\cdot m_{lz}^t + o_z, \qquad \forall t, \forall z, \quad (9)$$

where $N_L$ is the number of the neurons in the last hidden layer. Additionally, (10) and (11) represent the pre- and post-processors, discussed in Section II-A, respectively:

$$-\frac{2}{(\overline{X_{iz}} - \underline{X_{iz}})}X_{iz}^t + x_{iz}^t = -\frac{2\underline{X_{iz}}}{(\overline{X_{iz}} - \underline{X_{iz}})} - 1, \quad \forall i, \forall t, \forall z, \quad (10)$$

$$-\frac{(\overline{T_z} - \underline{T_z})}{2}y_z^t + T_z^t = \frac{(\overline{T_z} - \underline{T_z})}{2} + \underline{T_z}, \quad \forall t, \forall z, \quad (11)$$

where $\overline{X_{iz}}$ and $\underline{X_{iz}}$ are the maximum and minimum values, respectively, of the input training data $X_{iz}^t$. Moreover, $\overline{T_z}$ and $\underline{T_z}$ are the maximum and minimum values of the output training data $T_z^t$. Without loss of generality, (3)–(11) can be applied to different ANN architectures and activation functions.

## III. Supervised-learning-aided Optimal DR Scheduling of an HVAC System in a Multi-Zone Building

### A. Optimal DR via explicit ANN replication

The optimal price-based DR schedule of an HVAC system in a multi-zone building is determined using the explicit model (3)–(11) of the trained ANNs by solving

$$\underset{P_{set}^t \approx P^t}{\arg\min} \quad J_{DR} = \sum_{t=1}^{N_H} C_E^t P^t + \sum_{z=1}^{N_Z} \sum_{t=1}^{N_H} C_V^t \left( \Delta T_z^{tH} + \Delta T_z^{tL} \right) \quad (12)$$

subject to the following:

- Constraints on indoor temperatures $T_z^t$

$$T_z^t - \Delta T_z^{tH} \leq T_{z,\max}^t, \quad \forall t, \forall z, \quad (13)$$

$$-T_z^t - \Delta T_z^{tL} \leq -T_{z,\min}^t, \quad \forall t, \forall z, \quad (14)$$

$$HT_{z,\min}^t \leq T_z^t \leq HT_{z,\max}^t, \quad \forall t, \forall z, \quad (15)$$

- Constraints on the relationship between $T_z^t$ and $P^t$

$$n_{jz}^t - \sum_{i \in \mathbf{X}_{\mathbf{C}_z}}^{N_{C_z}} IW_{jiz} x_{iz}^t - \sum_{i \in \mathbf{X}_{\mathbf{F}_z}}^{N_{F_z}} IW_{jiz} x_{iz}^t = \quad (16)$$

$$\sum_{i \in \mathbf{X}_{\mathbf{E}_z}}^{N_{E_z}} IW_{jiz} x_{iz}^t + b_{jz}, \quad \forall j \in \mathbf{H}_{1z}, \forall t, \forall z,$$

$$-\sum_{j=1}^{N_J} HW_{kjz} m_{jz}^t + \sum_{s=1}^{N_S} q_{skz}^t = b_{kz} - r_o, \quad (17)$$

$$\forall k \in \mathbf{H}_{K(=J+1 \geq 2)z}, \forall t, \forall z,$$

$$X_{iz}^t - T_z^{(t-i \cdot \Delta t)} = 0, \quad \forall i \in \mathbf{X}_{\mathbf{F}_z}, \forall t \geq (i+1)\Delta t, \forall z, \quad (18)$$

$$X_{iz}^t = T_{z,pre}^{(N_H + (t-i\Delta t))}, \quad \forall i \in \mathbf{X}_{\mathbf{F}_z}, \forall t \leq i \cdot \Delta t, \forall z, \quad (19)$$

$$-1 \leq x_{iz}^t \leq 1, \quad \forall i \in \{\mathbf{X}_{\mathbf{C}_z}, \mathbf{X}_{\mathbf{F}_z}\}, \forall t, \forall z \quad (20)$$

$$-1 \leq y_z^t \leq 1, \quad \forall t, \forall z, \quad (21)$$

and (5)–(11),

- Constraints on time-delayed power inputs $P^{(t-\tau)}$

$$X_{iz}^t = P^{(t-(i-1)\Delta t)}, \quad \forall i \in \mathbf{X}_{\mathbf{C}_z}, \forall t, \forall z, \quad (22)$$

$$X_{iz}^{(t+(i-1)\Delta t)} - X_{(i+1)z}^{(t+i \cdot \Delta t)} = 0, \quad \forall i \in \mathbf{X}_{\mathbf{C}_z}, 1\,h \leq t < t_e, \forall z, \quad (23)$$

$$X_{iz}^t = 0, \quad \forall i \in \mathbf{X}_{\mathbf{C}_z}, \quad (24)$$

$$t \leq (i-1) \cdot \Delta t, \; N_H - (i-1) \cdot \Delta t \leq t \leq N_H, \forall z.$$

The objective function (12) aims to minimize the operating cost of the HVAC system: i.e., the 24-h sum of the hourly varying electricity prices $C_E^t$ multiplied by the power inputs $P^t$ of the HVAC system. Note that the optimal schedule of $P_{set}^t$ is determined as the optimization result for the direct control method, discussed in Section II. For the application of the indirect method, the argument includes $T_{z,set}^t$, along with additional constraints on $P^t \approx P_{set}^t = f(T_{z,set}^t, T_z^t)$. The remaining terms in (12) represent the penalties incurred when $T_z^t$ goes above the maximum limit $T_{z,max}^t$ by $\Delta T_z^{tH}$ or below the minimum limit $T_{z,min}^t$ by $\Delta T_z^{tL}$, as shown in (13) and (14), respectively. In other words, the temperature boundary conditions are relaxed by including $\Delta T_z^{tL}$ and $\Delta T_z^{tH}$ in (12), which allows the optimization problem to be reliably solved. Note that for each zone, different values of $T_{z,max}^t$ and $T_{z,min}^t$ can be set, for example, based on the thermal preferences of occupants. To prevent an excessive increase or decrease in $T_z^t$, (15) describes the hard boundary conditions.

The second set of constraints, (16)–(21), represents the relationship between the input variables, particularly $P^t$, and the output variable $T_z^t$ of the trained ANN for each zone, as discussed in Section II. Specifically, (3) can be equivalently expressed as (16), where $x_{iz}^t$ are divided into controllable, feedback, and environmental input variables (i.e., $P^t$, $T_z^{(t-1)}$, and $\mathbf{E}^t$ for $i \in \mathbf{X}_{\mathbf{C}_z}$, $i \in \mathbf{X}_{\mathbf{F}_z}$, and $i \in \mathbf{X}_{\mathbf{E}_z}$, respectively), as well as into the corresponding time-delayed inputs. Note that $n_{jz}^t$ in (16) is equivalent to the sum of $q_{sjz}^t$, as shown in (5-b). Analogously, (4) can be equivalently expressed as (17) using the linear expression of $n_{kz}^t$: i.e., $n_{kz}^t = r_0 + \Sigma_s q_{skz}^t$. Furthermore, (18) and (19) reflect the feedback loops of $T_z^t$. In particular, (18) shows the relationship between the feedback input variables $X_{iz}^t$ for $i \in \mathbf{X}_{\mathbf{F}_z}$ and the output variable $T_z^t$ of the ANNs for $t \geq (i+1) \cdot \Delta t$. The unit sampling time $\Delta t$ is set to 1 h to reflect the hourly varying $C_E^t$. For $t \leq i \cdot \Delta t$, $X_{iz}^t$ for $i \in \mathbf{X}_{\mathbf{F}_z}$ is set to $T_z^{(N_H + (t-i \cdot \Delta t))}$, as shown in (19), measured one day before the scheduling time period. In addition, (20) shows that the normalized values of the controllable and feedback input variables, obtained using (10), vary between –1 and 1; this also ensures $0 \leq P^t \leq P_{rated}$ for all $t$. Similarly, (21) represents the maximum and minimum values of $y_z^t$. Note that $y_z^t$ is linked to (17) via (5)–(9) and is reverse-transformed to $T_z^t$ in (13)–(15) using (11). Therefore, (5)–(11) are integrated into the second set of constraints.

To complete the integration of the trained ANNs into the optimization problem, (22) and (23) represent the constraints on the input neurons that receive the time-delayed power inputs of the HVAC system. Furthermore, it is assumed in (24) that the HVAC system turns off (i.e., $P^t = 0$) after working hours (i.e., $t \geq t_e$) when the building is relatively unoccupied. For pre-cooling, the HVAC system can operate (i.e., $P^t \geq 0$) during the early morning before people start to arrive for work.

The optimization problem (5)–(24) can be widely applied to practical buildings without significant modification; only the coefficients (i.e., $IW_{jiz}$, $HW_{kjz}$, $LW_{lz}$, $b_{j(k)z}$, and $o_z$) of the ANNs are expected to change according to the HVAC load characteristics and building thermal dynamics. Moreover, (5)–(24) include only linear equations and binary variables. Therefore, the optimization problem can be solved using an off-the-shelf MILP solver, ensuring global optimality of the DR schedule.

### B. Optimal DR via supervised-learning-aided meta-prediction

The EAR method can be further exploited to develop the SLAMP method that enables the direct determination, or more accurately predicticton, of the optimal solution to (5)–(24), considering the forecast of $C_E^t$ and $\mathbf{E}^t$ for $1 \leq t \leq N_H$. Specifically, as shown in Part 1 of Algorithm 1, (5)–(24) are formulated using the EAR method and iteratively solved offline, such that the optimal schedules of $P^t$ and $T_z^t$, as well as the corresponding $E_C = \Sigma_t C_E^t \cdot P^t$ and $T_V = \Sigma_z \Sigma_t C_V^t \cdot (\Delta T_z^{tH} + \Delta T_z^{tL})$, are obtained for historical or forecast data $\mathbf{m} = [t, C_E^t, \mathbf{E}^t]$. Fig. 6(a) shows that using the optimal $P^t$ and $T_z^t$, $\mathbf{m}$ is extended to $\mathbf{M} = [t, C_E^t, \mathbf{E}^t, T_{z,opt}^t, P_{opt}^t]$ and further to the input and output datasets $\mathbf{I}$ and $\mathbf{O}$, respectively, which are used for DNN training and testing in Part 2 of Algorithm 1. In particular, **I** also includes the time-delayed data of **M** to improve the training and testing performances of the DNN **N**. Moreover, as shown in Fig. 6(b), the datasets **I** and **O** are randomly shuffled, resulting in **D** = [**I**$_S$, **O**$_S$], such that the DNN training and testing datasets **D**$_r$⊂[**I**$_S$, **O**$_S$] and **D**$_e$⊂[**I**$_S$, **O**$_S$], respectively, include various daily profiles of $C_E^t$ and **E**$^t$, as well as the optimal schedules of $T_z^t$ and $P^t$, for $1 \le t \le N_H$. This random shuffling can mitigate DNN over-fitting to the profiles of $C_E^t$ and **E**$^t$ that are observed only during specific time periods (e.g., months or seasons).

In Part 2, the DNN **N** is trained and tested using **D**$_r$ and **D**$_e$, respectively, for every combination **W**($c$) of the network parameters that characterize the time delays $\tau$ of the input data **I**$_S$, the number of hidden layers $G$, the types of activation functions **F**, and the number of hidden neurons **U** in each layer. As in the case of the ANNs, the search range for the DNN architecture selection is determined by the minimum and maximum values of $\tau$, $G$, and **U** and the number of **F** types, as shown in Table III; the sigmoid functions or ReLUs are considered as the **F** types, similar to the case of the parameter set **Z** for the ANN architecture, as shown in Table II. If the search range is set too small, the DNN **N** will have the low performances during the training and testing, failing to accurately predict the optimal DR schedule of the HVAC system for the next 24 h. If the search range is too large, significant computational time will be required to select the parameter set **W** that results in high performances for the training and testing datasets. Because this paper focuses on developing and verifying the SLAMP method for the optimal day-ahead scheduling of the HVAC system operation, the search range has been set to be rather large, such that the selected DNN **N** can achieve the high training and testing performances, while being the least over-fitted.

**Algorithm 1.** SLAMP algorithm for optimal DR scheduling

1 : **Input:** **m** = [$t$, $C_E^t$, **E**$^t$]
 *Part 1. Datasets **D**$_r$ and **D**$_e$ for training and testing of the DNN.*
2 : **Repeat** until the end day for dataset **m**:
3 :  Solve (5)–(24) → [$P^t$, $T_z^t$]$_{opt}$
4 :  Extend **m** to **M** = [$t$, $C_E^t$, **E**$^t$, $T_{z,opt}^t$, $P_{opt}^t$]
5 :  $E_C = \Sigma_t C_E^t \cdot P^t$, $T_V = \Sigma_z\Sigma_t C_V^j \cdot (\Delta T_z^{tH} + \Delta T_z^{tL})$ → [$E_C$, $T_V$]$_{opt}$
6 : Process the optimal dataset **M**:
7 :  Set **I**($t$,:) = [$t$, $C_E^{(t-\tau1)}$, **E**$^{(t-\tau2)}$, **T**$_{z,opt}^{(t-\tau3)}$, $P_{opt}^{(t-1-\tau4)}$] and **O**($t$,:) = $P_{opt}^t$
8 :  Randomly shuffle the daily profiles of **I** and **O** (see Fig. 6(b)) as:
9 :   e.g., **I**$_S$ = [**I**(**t**$_{15}$,:); ⋯ ; **I**(**t**$_4$,:)] and **O**$_S$ = [**O**(**t**$_{15}$,:); ⋯ ; **O**(**t**$_4$,:)]
10 :    where **t**$_d$ = [1 h; ⋯ ; 24 h] for a day $d$
11 :  Define **D**$_r$, **D**$_e$ ⊂ **D** = [**I**$_S$, **O**$_S$] as the training and testing datasets.
 *Part 2. DNN **N**$^*$ affording the best predictive performance.*
12 : Define search ranges for the network parameter set **W** = [$\tau$, $G$, **U**, **F**]:
13 :  time delay ($\tau$) : $0 \le \tau_1, \tau_2, \tau_3, \tau_4 \le \tau_{1,max}, \tau_{2,max}, \tau_{3,max}, \tau_{4,max}$
14 :  no. of hidden layers: $1 \le G \le G_{max}$, activation functions (**F**): $F_1, \cdots, F_G$
15 :  no. of hidden neurons (**U**): $U_{1,min},\cdots,U_{G,min} \le U_1,\cdots,U_G \le U_{1,max},\cdots,U_{G,max}$
16 : **Repeat** until the final combination of **W** is attained:
17 :  Initialize DNN architecture **N** with **W**($c$) ← **W**
18 :  Train **N** with **D**$_r$ → [$P^{t\prime}$, $T_z^{t\prime}$]$_{opt}$ → $e_{pr}=e$(**P**, **P**$\prime$), $e_{tr}=e$(**T**$_z$, **T**$_z\prime$)
19 :  Test **N** with **D**$_e$ → [$P^{t\prime}$, $T_z^{t\prime}$]$_{opt}$ → $e_{pe}= e$(**P**, **P**$\prime$), $e_{te}=e$(**T**$_z$, **T**$_z\prime$)
20 :   **Repeat** till the end day for **D**$_e$
21 :    $E_C\prime = \Sigma_t C_E^t \cdot P^{t\prime}$, $T_V\prime = \Sigma_z\Sigma_t C_V^j \cdot (\Delta T_z^{tH} + \Delta T_z^{tL})$ → [$E_C\prime$, $T_V\prime$]$_{opt}$
22 :  $e_c = \omega_1 \cdot e_{pr} + \omega_2 \cdot e_{tr} + \omega_3 \cdot e_{pe} + \omega_4 \cdot e_{te} + \omega_5 \cdot e$(**E**$_C$, **E**$_C\prime$)+$\omega_6 \cdot e$(**T**$_V$, **T**$_V\prime$)
23 :  If $e_c > e^*$, then $e^* \leftarrow e_c$, **W**$^* \leftarrow$ **W**($c$), and **N**$^* \leftarrow$ **N** with **W**$^*$
24 : **Output**: [$P^{t\prime}$, $T_z^{t\prime}$]$_{opt}$ of **N**$^*$ for $C_E^t$ and **E**$^t$ during the next scheduling day.

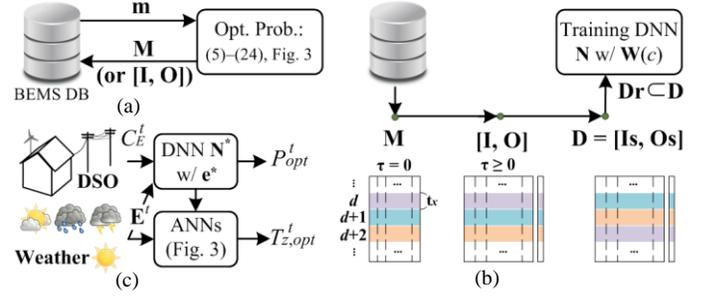

Fig. 6. (a) Dataset **M** including the optimal solutions of (5)–(24); (b) training dataset **D**$_r$ of the deep neural network (DNN); and (c) direct determination of the optimal DR schedule using the selected DNN **N**$^*$.

TABLE III. SEARCH RANGE FOR THE PARAMETER SET **W** OF THE DNN

| Parameter | $\tau_{1,max}$ | $\tau_{2,max}$ | $\tau_{3,max}$ | $\tau_{4,max}$ | $\Delta\tau_1$ | $\Delta\tau_2$ | $\Delta\tau_3$ | $\Delta\tau_4$ | $G_{max}$ | $\Delta G$ | $U_{g,max}$ | $U_{g,min}$ | $\Delta U_g$ |
|---|---|---|---|---|---|---|---|---|---|---|---|---|---|
| Value | 4 | 4 | 4 | 4 | 1 | 1 | 1 | 1 | 10 | 1 | 20 | 5 | 5 |

Given $C_E^t$ and **E**$^t$, the optimal schedule of $P^t$ can be meta-predicted using the DNN **N** trained with **W**($c$). The corresponding $T_z^t$ also can be obtained using the trained ANNs, as shown in Fig. 3. The DNN performance is evaluated using the weighted sum $e_c$ of the NMSEs [i.e., (1)] of $P^t$, $T_z^t$, $E_C$, and $T_V$ for all daily profiles in **D**$_r$ and **D**$_e$. The parameter set **W**($c$) leading to the highest value of $e^*$ is selected as **W**$^*$ to implement the DNN **N**$^*$, and consequently to determine the optimal DR schedules of $P^t$ and $T_z^t$ for $C_E^t$ and **E**$^t$ during the next scheduling time period, as shown in Fig. 6(c).

The SLAMP method makes it possible for the DSO or DR aggregators to predict the optimal response of the HVAC system to variations in electricity pricing for the time-varying thermal environment of the multi-zone building. Consequently, this lessens the necessity for the DR participants to solve the optimal scheduling problem (5)–(24) on a daily basis, particularly after sufficient data on the solutions to (5)–(24) have been collected. In addition, the computational time required for optimal DR scheduling is significantly reduced, as discussed in Section IV-D, because the optimal schedule can be obtained directly as the outputs of the trained DNN, rather than by solving (5)–(24). The SLAMP method is developed using the EAR method based on building operating data, rather than using physics-based models. Therefore, the SLAMP method is expected to effectively assist the DSO in identifying the optimal price sensitivities of actual, individual HVAC systems and determining the optimal RTP rates $C_E^t$ to induce the optimal HVAC load demands at different buses in a large-scale distribution grid for the time-varying conditions of building thermal environments. In other words, it can replace conventional price-and-quantity curves that are very simplified [21]. This paper focuses on the development and demonstration of the SLAMP method using the EAR method, and its applications are not further discussed here.

## IV. CASE STUDIES AND SIMULATION RESULTS

### A. Test building and simulation conditions

The proposed SL-based DR strategy was tested for a multi-zone building with the HVAC system, as shown in Fig. 7. The test building was modeled based on the DOE commercial

-7-

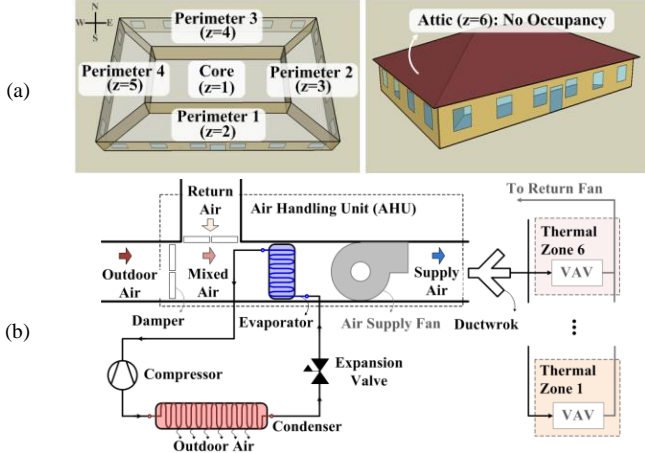

Fig. 7. Models of (a) a multi-zone office building and (b) an HVAC system used to test the proposed ML-based DR strategy.

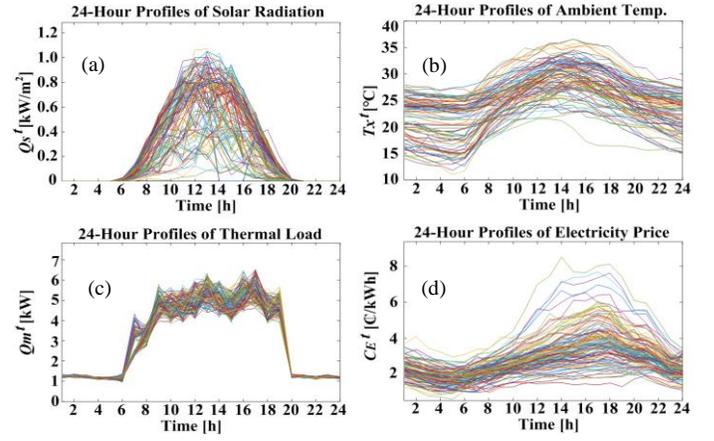

Fig. 8. Test conditions from May to September in 2012: (a) solar insolation $Q_s^t$, (b) ambient temperature $T_x^t$, (c) building thermal loads $Q_m^t$, and (e) electricity prices $C_E^t$. The test conditions in 2013 were similarly represented.

building benchmark for a small office [36], [37]. Specifically, the building has six thermal zones, including an attic that is unoccupied. The perimeter zones directly face the outside environment and experience variations in solar insolation at different time periods of the day. The core zone has the largest area, where the temperature is affected by large internal heat gains from occupants and external heat gains delivered through the perimeter zones via heat conduction and convection. The HVAC system has the rated power of $P_{rated} = 30.0$ kW [5] and provides the cooling rate $Q^t$ to control the temperatures in the core and perimeter zones within acceptable ranges. It was assumed that the occupants in zones $z = 1, 3$, and 4 feel comfortable for $T_{z,min}^t = 20°C \leq T_z^t \leq T_{z,max}^t = 25°C$ during 8 h $\leq t \leq$ 19 h, and that those in zone $z = 2$ and 5 are satisfied with $T_z^t$ between 19°C and 23°C over this time period.

The multi-zone building was simulated using historical weather data from April 1 to September 30 in the years 2012 and 2013 in Boston, MA [38] to establish the training and testing data for the case studies. The total number of historical datasets $[t, C_E^t, \mathbf{E}^t, T_z^t, P^t]$ used for the case studies were estimated as 8784 and 13 with respect to time and objects, respectively; note that the time-delayed data for the objects $C_E^t$, $\mathbf{E}^t$, $T_z^t$, and $P^t$ were not considered in the estimation. The network training processes, discussed in Sections II and III, were performed using approximately 80% of the 8784 datasets for each thermal zone in the building. The remaining data were used to test the performances of the trained networks, and consequently the proposed SL-based DR scheduling strategy using the EAR and SLAMP methods.

Fig. 8(a) and (b) show the 24-h profiles of the 1-h average solar insolation $Q_s^t$ and outdoor temperature $T_x^t$, respectively. For these weather conditions, $T_z^t$ data were obtained using the physics-based model of the test building provided in [5] and [37], where the profiles of thermal energy load $Q_m^t$ could be represented as shown in Fig. 8(c) based on the operation of the real commercial building [31], [39]. The thermal loads are increased at $t = 7$ h or 8 h and maintained at high levels until $t = 19$ h, when people start arriving at work and then leave the building, respectively. Fig. 8(d) shows the profiles of the electricity prices [40]; the price remained low for early morning, increased over time when the thermal loads were maintained at high levels, and then became lower again for $t \geq 19$ h.

Case studies were then conducted to compare the proposed SL-based strategy with the conventional physics-based strategy (i.e., the proposed EAR and SLAMP methods with the fully informed physics-model-based method) with respect to operating cost and computational time. The physics-based strategy can be considered as an ideal strategy, because it was developed based on the ideal assumption that DR aggregators or end-users are informed of all parameters, listed in Table IV, to model the HVAC system operation and building thermal environments; this assumption is not valid for most buildings in practice. Therefore, the comparison with the ideal strategy, rather than other ML-based methods, enables the conservative evaluation of the performances of the proposed SL-based methods; in [7]–[17], the optimal DR schedules were compared with non-DR profiles currently observed or pre-determined by fixed temperature set-points. In Section IV, the proposed DR methods were also compared with the non-DR method.

TABLE IV. BUILDING DATA AND MODELING PARAMETERS FOR THE PROPOSED SL-BASED AND CONVENTIONAL PHYSICS-MODEL-BASED DR STRATEGIES

| Parameters | Explanations |
|---|---|
| 1. Data for both proposed and conventional DR strategies | |
| $P^t$ | input power of the HVAC system at time $t$ |
| $T_x^t, T_g^t, T_z^t$ | ambient, ground, and zone temperatures at time $t$ |
| $Q_s^t, Q_m^t$ | solar radiation and internal thermal load at time $t$ |
| 2. Additional modeling parameters for the conventional DR strategy | |
| 2.1 *HVAC system operation* | |
| $f_c^t$ | compressor frequency at time $t$ |
| $T_e^t$ | evaporator (mixed) air temperature at time $t$ |
| $k_f^t, k_x^t, k_e^t, k_o^t$ | data-driven coefficients to estimate the input power $P^t$ of the HVAC system for $f_c^t, T_x^t, T_e^t$, and offset power |
| $q_f^t, q_x^t, q_e^t, q_o^t$ | data-driven coefficients to estimate cooling rate $Q^t$ of the HVAC system for $f_c^t, T_x^t, T_e^t$, and offset cooling rate |
| 2.2 *Building thermal environment* | |
| $Q_z^t$ | cooling rate supplied by the HVAC system to zone $z$ at time $t$ |
| $Q_{cz}^t, Q_{rz}^t$ | zonal convective and radiative heat gains in zone $z$ at time $t$ |
| $Q_{wz}^t, Q_{sz}^t$ | window-transmitted and wall-incident solar heat gains in zone $z$ at time $t$ |
| $a_{mn}^t$ | data-driven coefficient for the effect of $T_z^t$ for $z = n$ on $T_z^t$ for $z = m$ |
| $b_z^t, c_z^t, d_z^t, e_z^t, f_z^t, g_z^t$ | data-driven coefficient for $T_x^t, Q_{rz}^t, Q_{cz}^t, Q_{wz}^t, Q_{sz}^t$, and $T_g^t$, respectively |
| $N_r, N_{oz}^t$ | number of rooms in zone $z$ and occupants in zone $z$ at time $t$ |
| $T_{zf}^t$ | zone temperature at time $t$ under no cooling condition for the period from 1 to $t$ |
| $F_{zu}^{t\tau}$ | linear gradient of $T_z^t$ at time $t$ resulting from input power block $u$ of the HVAC system at time $\tau$ |
| $P_u$ | $u$th boundary when the rated power of the HVAC system is divided into piecewise linear blocks |
| $t_d$ | maximum time delay within which $T_z^t$ is expressed using the inverse transfer function of $\mathbf{E}^t$ |

### B. ANNs for modeling of building thermal dynamics

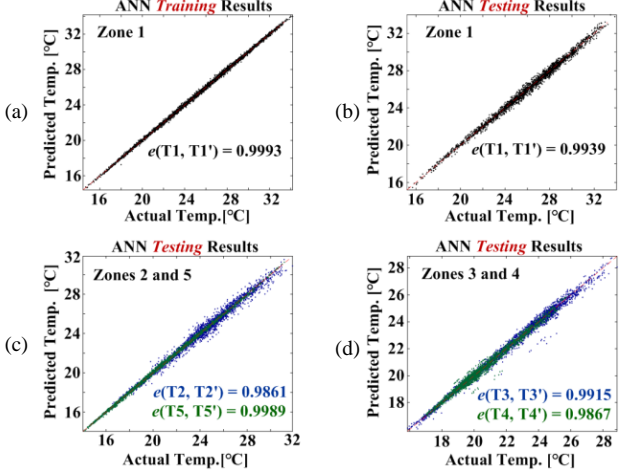

Fig. 9. Comparisons of the actual and estimated indoor temperatures: (a), (b) the ANN training and test results, respectively, for $z = 1$; and (c), (d) the ANN test results for $z = 2$ and 5 and $z = 3$ and 4, respectively.

TABLE V. NMSEs OF THE ANNs FOR THE EAR METHOD

| NMSEs | Zone 1 | Zone 2 | Zone 3 | Zone 4 | Zone 5 |
|---|---|---|---|---|---|
| (1) Training | 0.9993 | 0.9994 | 0.9994 | 0.9998 | 0.9999 |
| (2) Testing | 0.9939 | 0.9861 | 0.9915 | 0.9867 | 0.9989 |
| {(1)–(2)}/(1) [%] | 0.5404 | 1.3308 | 0.7905 | 1.3103 | 0.1000 |

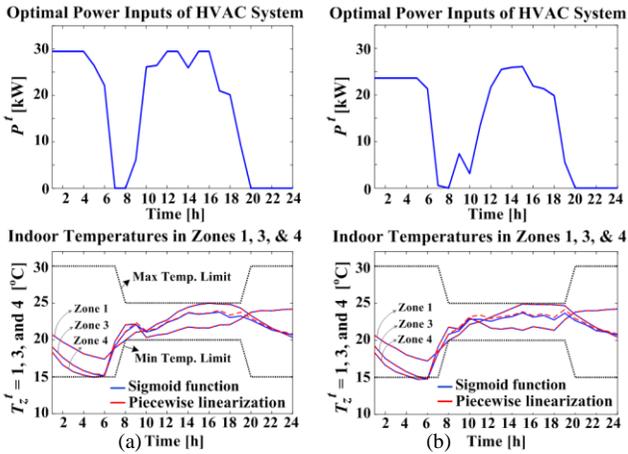

Fig. 10. Comparisons between the zone temperatures estimated from the trained ANNs using the sigmoid activation functions and the corresponding piecewise linearized equations for the optimal $P^t$ schedules.

Fig. 9 and Table V show the training and test results of the ANNs for the EAR method, discussed in Sections II and III-A, where the ANNs reflect the thermal dynamics of Zones 1–5 in the test building: i.e., $T_z^t$ for $P^t$ and $\mathbf{E}^t$. Testing was conducted with the feedback loops of the ANNs in a closed state, allowing $T_z^{(t-\tau)\prime}$ to be used to predict $T_z^t$, whereas training was performed with open feedback loops so that the actual data for $T_z^{(t-\tau)}$ could be fed into the ANNs. This implies that the test results in Fig. 9 and Table V show the performances of the trained ANNs used for the 24-h scheduling. Specifically, in Fig. 9, the $x$- and $y$-axes represent $T_z^t$ obtained from the building model and $T_z^{t\prime}$ estimated from the ANNs, respectively. In Fig. 9(a) and (b), $e(\mathbf{T}_{z=1}, \mathbf{T}\prime_{z=1})$ are estimated to be 0.9993 and 0.9939, respectively. Fig. 9(c) and (d) show the test results of the ANNs for $z = 2$ and 5 and $z = 3$ and 4, respectively. The corresponding $e(\mathbf{T}_z, \mathbf{T}_z\prime)$ are also estimated to be high, implying that the ANNs successfully reflect the complicated thermal dynamics of the multi-zone building and, consequently, can be used effectively in (5)–(24) to determine the optimal DR schedule of the HVAC system, while ensuring the thermal comfort of occupants.

In addition, Fig. 10 compares $T_z^t$ estimated from the ANNs using the sigmoid activation functions and the corresponding piecewise linear equations for $N_S = 5$. It shows the good precision of the piecewise linearization for the different profiles of the optimal $P^t$, consequently validating the application of the EAR method to the optimal DR scheduling.

### C. DNN for SLAMP-based optimal DR scheduling

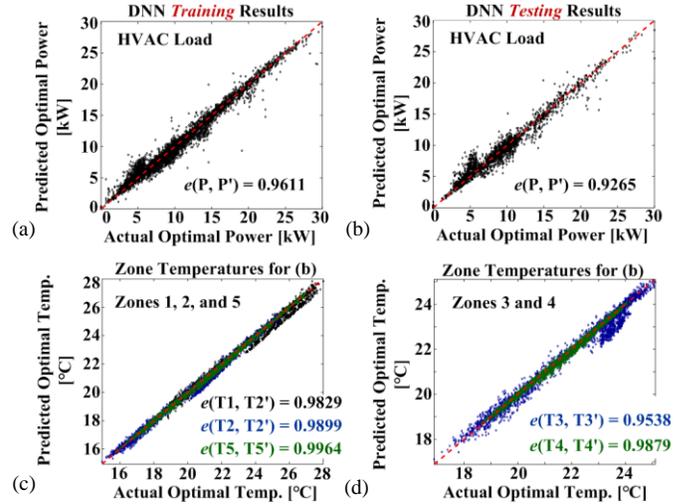

Fig. 11. Comparisons of the optimal power inputs of the HVAC system determined using the EAR and SLAMP methods: (a), (b) the DNN training and test results, respectively; and (c), (d) the corresponding zone temperatures for the test results shown in (b).

TABLE VI. NMSEs OF THE DNN AND ANNs FOR THE SLAMP METHOD

| NMSEs | Power | Zone 1 | Zone 2 | Zone 3 | Zone 4 | Zone 5 |
|---|---|---|---|---|---|---|
| (1) Training | 0.9611 | 0.9888 | 0.9936 | 0.9707 | 0.9947 | 0.9979 |
| (2) Test | 0.9265 | 0.9829 | 0.9899 | 0.9538 | 0.9879 | 0.9964 |
| {(1)–(2)}/(1) [%] | 3.6000 | 0.5907 | 0.3724 | 1.7410 | 0.6836 | 0.1503 |

Fig. 11 and Table VI show the training and testing results of the DNN for the SLAMP method, integrated with the EAR method, as discussed in Section III-B. The DNN reflects the optimal operation of the HVAC system for the time-varying electricity prices and building thermal conditions: i.e., $P^t_{opt}$ for $C_E^t$ and $\mathbf{E}^t$. As in Section IV-B, the trained DNN was tested with the feedback loops of $P^t$ in a closed state to estimate the performance of the multi-hour optimal scheduling. Specifically, the $x$-axis in Fig. 11(a) and (b) represents the optimal $P^t$ determined by solving (5)–(24), which was formulated using the explicit ANN models. The $y$-axis shows the optimal $P^{t\prime}$ directly estimated using the DNN, which was trained with historical data on the optimal solutions to (5)–(24) for various profiles of $\mathbf{m} = [C_E^t, \mathbf{E}^t]$. In Fig. 11(a) and (b), the NMSEs $e(\mathbf{P},$

-9-

**P′**) were calculated to be 0.9611 and 0.9265, respectively. Meta-prediction errors were often observed during the early morning hours, because the time-delayed data of $P^t$ have zero values for this time period; the HVAC system remained off for $20\ h \leq t \leq 24\ h$ one day before the scheduling day. This implies that the DNN performance can be further improved when $P^t$ is scheduled using a unit sampling time $\Delta t$ of less than 1 h, although this requires a larger volume of building and weather data with higher granularity. Fig. 11(c) and (d) compare the optimal $T_z^t$ and $T_z^{t\prime}$ estimated for the optimal $P^t$ and $P^{t\prime}$, respectively, in Fig. 11(b). It can be seen in Fig. 11 and Table VI that $e(\mathbf{T_z}, \mathbf{T_z'})$ were higher than $e(\mathbf{P}, \mathbf{P'})$, due to the large thermal capacity of the building. The minor error in the optimal $P^t$ marginally affects $T_z^t$, supporting the effectiveness of the SLAMP method.

*D. Comparison of the proposed EAR and SLAMP methods with the fully informed physics-model-based method*

As discussed in Section IV-A, the optimal schedules of the HVAC system operation were compared for three cases: i.e., the proposed EAR and SLAMP methods (Cases 1 and 2, respectively) and the conventional fully informed physics-model-based method (Case 3). In Case 3, the nonlinear variation in $T_z^t$ for the change in $P^t$ was approximated using a piecewise linear model [5], assuming that the complete information on the building thermal dynamics was available.

Figs. 12 and 13 show the optimal $P^t$ and $Q^t$ schedules of the HVAC unit and the corresponding $T_z^t$ for the profiles of the electricity price and building thermal conditions (i.e., $C_E^t$, $Q_m^t$, $Q_s^t$, and $T_x^t$) shown in Figs. 12(a), 12(b), and 13(a). Specifically, in Case 1, relatively high $P^t$ were scheduled for $1\ h \leq t \leq 6\ h$, whereas the heat gains from $Q_m^t$, $Q_s^t$ and $T_x^t$ were maintained at high levels for $8\ h \leq t \leq 19\ h$. As $C_E^t$ began to increase, lower $P^t$ were then scheduled for $7\ h \leq t \leq 11\ h$. In other words, due to the low energy price in the early morning, the HVAC load was shifted away from on-peak hours to run during off-peak hours, leading to the pre-cooling operation. This also enables efficient attainment of the cooling rate $Q^t$, because the building thermal environment in the early morning is suitable for HVAC system operation. The coefficients of performance (COPs) of the HVAC system were higher during the pre-cooling time period than the on-peak hours (i.e., $13\ h \leq t \leq 17\ h$), as can be seen by comparing the ratios of $Q^t$ to $P^t$. The pre-cooled temperatures gradually increased to almost $T_{z,max}^t$ and a little higher than $T_{z,max}^t$ for $15\ h \leq t \leq 18\ h$ in Fig. 13(a) and (b), respectively, mainly due to increases in $T_x^t$ and $Q_s^t$. To maintain $T_z^t$ between $T_{z,min}^t$ and $T_{z,max}^t$, the HVAC system was scheduled to operate with relatively high $P^t$ for $13\ h \leq t \leq 17\ h$. The HVAC unit then consumed less power for $18\ h \leq t \leq 19\ h$, as $T_x^t$ and $Q_s^t$ decreased during this time period. As listed in Table VII, $E_C$ for Case 1 was estimated at \$11.23, which is 11.47% less than \$12.69 for the non-DR case, where $P^t$ was maintained constant to control the time-average value of $T_z^t$ during $8\ h \leq t \leq 19\ h$ almost at the mid-point of $T_{z,max}^t$ and $T_{z,min}^t$. In addition, $E_C$ for Case 1 was only about 2.9% higher than that for Case 3, where the fully informed physics-based model of the building thermal dynamics enabled the HVAC system to operate with sharp variations in $P^t$ during the pre-cooling time period.

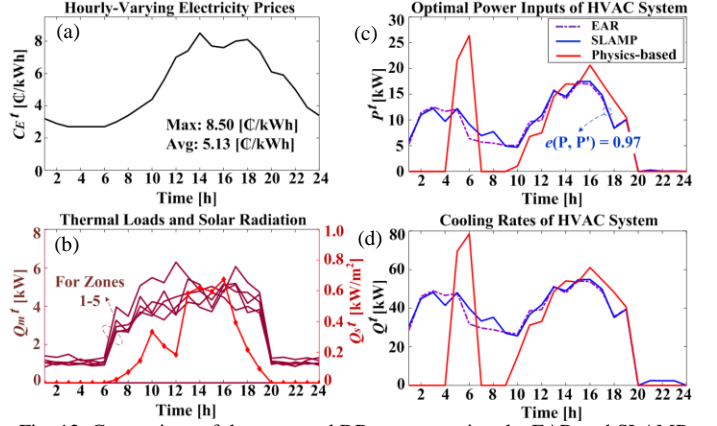

Fig. 12. Comparison of the proposed DR strategy using the EAR and SLAMP methods (Cases 1 and 2, respectively) with the physics-based strategy (Case 3): (a) electricity price $C_E^t$, (b) building thermal loads $Q_m^t$ and solar insolation $Q_s^t$, and (c), (d) optimal HVAC power input $P^t$ and cooling rate $Q^t$.

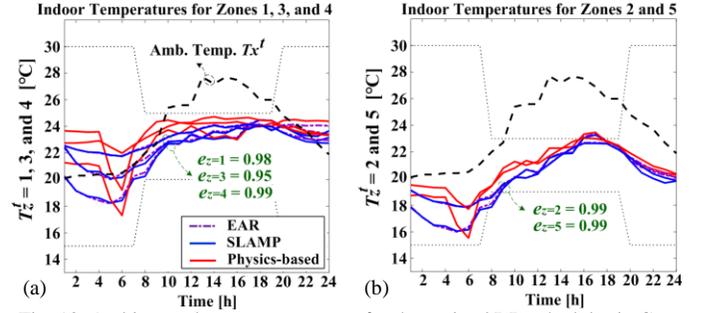

Fig. 13. Ambient and zone temperatures for the optimal DR schedules in Cases 1–3: (a) $T_z^t$ for $z$ = 1, 3, and 4 and (b) $T_z^t$ for $z$ = 2 and 5.

TABLE VII. COMPARISONS OF THE ENERGY COSTS FOR THE PROPOSED DR, FULLY INFORMED DR, AND NON-DR STRATEGIES: PROFILE 1 OF $C_E^t$ AND $\mathbf{E}^t$

| Profile 1 (Fig. 12) | Non-DR | Proposed DR | | Fully informed model-based DR |
|---|---|---|---|---|
| | | EAR | SLAMP | |
| Energy cost $E_C$ [\$] | 12.69 | 11.23 | 11.52 | 10.91 |
| Reduction rate [%] | - | 11.47 | 9.23 | 14.03 |

TABLE VIII. COMPARISONS OF THE COMPUTATION TIME FOR THE PROPOSED AND FULLY INFORMED DR STRATEGIES

| Computation time | | (1) Fully informed model-based DR | (2) Proposed DR | |
|---|---|---|---|---|
| | | | EAR | SLAMP |
| No. of binary decision variables | | 57 | 746 | - |
| Profile 1 (Fig. 12) | Computation time [s] | 87.98 | 775.65 | 0.26 |
| | Relative ratio, (2)/(1) | - | 8.8 | 3.0×10$^{-3}$ |
| Profile 2 (Fig. 14) | Computation time [s] | 76.24 | 757.54 | 0.25 |
| | Relative ratio, (2)/(1) | - | 9.9 | 3.3×10$^{-3}$ |

TABLE IX. COMPARISONS OF THE ENERGY COSTS FOR THE PROPOSED DR, FULLY INFORMED DR, AND NON-DR STRATEGIES: PROFILE 2 OF $C_E^t$ AND $\mathbf{E}^t$

| Profile 2 (Fig. 14) | Non-DR | Proposed DR | | Fully informed model-based DR |
|---|---|---|---|---|
| | | EAR | SLAMP | |
| Energy cost $E_C$ [\$] | 9.29 | 7.63 | 7.94 | 7.29 |
| Reduction rate [%] | - | 17.79 | 14.48 | 21.53 |

It can be seen in Figs. 12 and 13 that for Case 2, the optimal $P^t$ and corresponding $T_z^t$ were very similar to those for Case 1, demonstrating the good performance of the SLAMP method. The slight differences between the optimal $P^t$ profiles for Cases 1 and 2 were observed particularly during $t \leq 8\ h$. The HVAC system remained off for $19\ h \leq t \leq 24\ h$ one day before the scheduling day. The zero values of $P^{(t-1-\tau)}$ in the input dataset $\mathbf{I}$ in Algorithm 1 contributed little to the meta-prediction of the variations in $P^t$ for $t \leq 8\ h$ after the rather abrupt increase at $t = 1$.

The difference remained small for $t \geq 9$ h, when all $P^{(t-1-\tau)}$ had non-zero values. The differences in $P^t$ caused imperceptible changes in $T_z^t$, as shown in Fig. 13, given the building thermal capacity. As listed in Table VIII, the computational time for Case 2 was 0.26 s, whereas for Case 3, it was estimated as 81.90 s. The simulations were performed on a computuer with a four-core, 3.5-GHz CPU and 32-G RAM. The simulation results verified the effectiveness of the SLAMP method in the optimal DR of the HVAC system, being practical and affordable in terms of operating costs and computational time.

Figs. 14 and 15 show the optimal DR scheduling results for different profiles of $C_E^t$ and $\mathbf{E}^t$. For Cases 1 and 2, the HVAC system still operated to exploit the large thermal capacity of the pre-cooled building. In Fig. 14(a) and (b), $C_E^t$ differed less between the on- and off-peak hours and $Q_s^t$ was maintained higher, compared with those shown in Fig. 12(a) and (b), respectively. Therefore, the HVAC load was less shifted, resulting in the relatively larger difference in $P^t$ over time. This finding indicates that the proposed SL-based strategy successfully reflects the load shifting or curtailment capabilities of the HVAC system in response to the time-varying electricity prices and building thermal conditions. As listed in Table IX, $E_C$ for Case 1 was 17.79% less than that for the non-DR case and only approximately 4.7% higher than that for Case 3. For the profiles of $C_E^t$ and $\mathbf{E}^t$, $P^t$ and $T_z^t$ for Case 2 were still very similar to those for Case 1, demonstrating the performance of the SLAMP method. As discussed above, the slight differences in $P^t$ during $t \leq 6$ h were attributed to the zero power inputs of the HVAC system for 19 h $\leq t \leq$ 24 h one day before the scheduling day. Fig. 15 shows that $T_z^t$ was negligibly affected by the differences due to the building thermal capacity. The computational time for Case 2 was 99.71% less (i.e., from 76.24 s to 0.25 s) than that for Case 3, as shown in Table VIII.

*E. Discussion*

The case study results show the effectiveness of the proposed EAR and SLAMP methods. The operating costs were reduced to a level comparable to those for the fully informed model-based method, while the computational time for optimal DR scheduling was significantly reduced. The HVAC load was effectively shifted from on-peak to off-peak hours in response to the variations in the RTP rates and building thermal conditions, while ensuring the thermal comfort of occupants.

The more training data available for various conditions related to the HVAC system operation and building thermal environment, the better generalization of the network training, the wider search space for the optimal solution to (5)–(24), and the further reduction in operating costs that can be achieved. Dependence on training data is a common issue with SL. Due to the development of IoT sensing and big data technologies, more building operating data can be collected, enabling ANNs to be more robust to over-fitting and consequently further improving the applicability and performance of SL-based DR strategies. Because it takes an extended period of time to collect large quantities of operating data, the performances of SL-based strategies, including the proposed methods and those discussed in [10]–[17], are expected to be low, particularly for a new building, for which the size and variability of building operating data are limited. Note that the performances of RL-based or physics-based strategies are also dependent on operating data, which are essential for data-driven modeling of HVAC system operations and building thermal dynamics. Therefore, the immediate application of these strategies to a new building remains as a topic for future research. Several methods for overcoming the challenges have been discussed in recent papers (e.g., [41] and [42]), including online learning using online data estimation and coordination with physics-model-based methods. Using these methods, the SL-based DR strategies can be initiated and then gradually improved as the size and variability of the building operating data continue to increase.

Figs. 12–15 show the optimal schedules of $P^t$ and $T_z^t$ using the day-ahead forecast of $C_E^t$ and $\mathbf{E}^t$. A number of sophisticated forecast models were developed in conjunction with techniques to mitigate over-fitting [33]–[35]; e.g., applying weight decay and dropout regularization, stopping the training early, increasing the size of training data, and changing the network (or tree) architectures. The proposed DR strategy can also be implemented considering real-time uncertainties in $C_E^t$ and $\mathbf{E}^t$. For implementation, the difference between the forecast results of the long-term (i.e., day-ahead) and short-term (i.e., hour- or minute-ahead) models needs to be estimated online. The neural networks for the EAR and SLAMP methods can then be re-trained online to reflect the forecast errors and re-schedule the optimal $P^t$ and $T_z^t$.

## V. CONCLUSION

This paper proposes a new SL-based strategy for the optimal DR of an HVAC system in a multi-zone building. The ANNs were trained using an SL algorithm to reflect the complicated thermal dynamics of the multi-zone building. The ANNs were then replicated using piecewise linear equations and explicitly integrated into the optimal scheduling problem. Using an off-the-shelf MILP solver, the optimal solutions were obtained for historical BEMS data on the electricity prices and building thermal conditions. In the SLAMP method, the DNN was trained with these optimal solutions to directly determine the optimal power input schedule of the HVAC system for the next 24 h. The simulation studies were conducted for three cases: i.e., the proposed SL-based DR strategy using the EAR and SLAMP methods, and the fully informed physics-model-based strategy. The case study results demonstrated the effectiveness of the proposed EAR and SLAMP methods with respect to practical applicability, operating cost, and computational time. The simulations also verified that the EAR and SLAMP methods retain the load-shifting capability of the HVAC system in response to time-varying electricity prices and ensure the thermal comfort of the occupants.

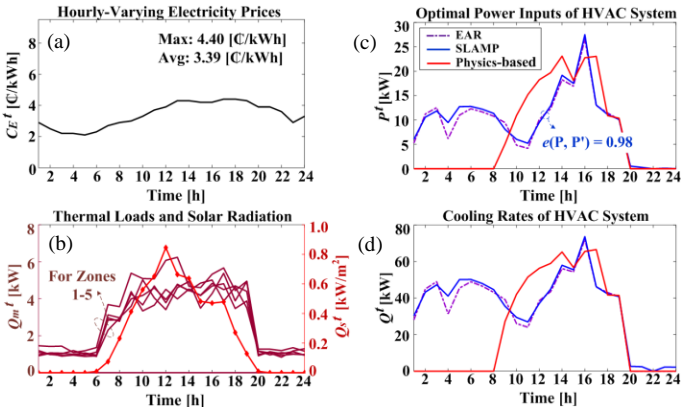

Fig. 14. Comparison of the proposed DR strategy using the EAR and SLAMP methods (Cases 1 and 2, respectively) with the physics-based strategy (Case 3): (a) $C_E^t$, (b) $Q_m^t$ and $Q_s^t$, and (c), (d) optimal $P^t$ and $Q^t$.

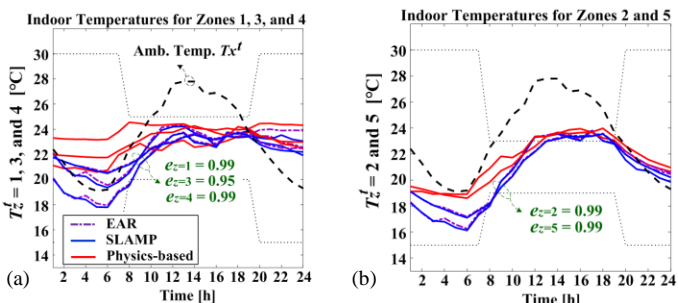

Fig. 15. Zone temperatures for the optimal DR schedules in Cases 1–3.

Further work is required with regard to the application of the proposed DR strategy to new or non-retrofit buildings, for which the size and variability of the building operating data are limited. It will be of interest to develop a procedure that ensures continuous improvement in the performance of the DR strategy as training data continue to be collected. Coordination with RL-based or physics-based strategies will also be considered in the procedure. It is also important to adopt different ML models. For example, support vector machines and random decision forests can be robust to over-fitting, which will further improve the performance of the proposed strategy. Another area for future research includes developing a cooperative strategy of multiple HVAC units in a large-scale distribution grid, which can be effectively achieved using the EAR and SLAMP methods. Cooperation will significantly increase the demand-side flexibility provided to the grid; however, the application of SL has rarely been considered in previous studies.